\documentclass{llncs}
\begin{document}

\title{Extending Automated Deduction for Commonsense Reasoning}

\author{Tanel Tammet}

\authorrunning{T. Tammet.}
%
\institute{Tallinn University of Technology, Tallinn, Estonia\\
\email{tanel.tammet@taltech.ee}}
\maketitle              
\begin{abstract}
Commonsense reasoning has long been considered as one of the holy grails of 
artificial intelligence. Most of the recent progress in the field has been achieved
by novel machine learning algorithms for natural language processing.
However, without incorporating logical reasoning, these algorithms remain 
arguably shallow. With some notable exceptions, developers of practical 
automated logic-based reasoners have mostly avoided focusing on the problem. 
The paper argues that the methods and algorithms used by existing automated reasoners
for classical first-order logic can be extended towards commonsense reasoning. 
Instead of devising new specialized logics we propose a framework of extensions
to the mainstream resolution-based search methods to make these capable of performing
search tasks for practical commonsense reasoning with reasonable efficiency. 
The proposed extensions mostly rely on operating on ordinary proof trees and 
are devised to handle commonsense knowledge bases containing inconsistencies, 
default rules, taxonomies, topics, relevance, confidence and similarity measures.
We claim that machine learning is best suited for the construction of 
commonsense knowledge bases while the extended logic-based methods would 
be well-suited for actually answering queries from these knowledge bases.

\end{abstract}
\section{Introduction}

Ability to perform ``commonsense reasoning'' (CSR) has been considered one of the big goals of 
Artificial Intelligence (AI) for a very long time. One of the first papers \cite{ref_mccarthy_1} introducing 
the term comes from John McCarthy already at 1958. 
The exact meaning of CSR is -- perhaps necessarily so -- vague.
Wikipedia \cite{ref_wikipedia_1} states, following \cite{ref_Davis_Marcus_1}: ``Commonsense reasoning is 
one of the branches of artificial intelligence (AI) 
that is concerned with simulating the human ability to make presumptions about the type and 
essence of ordinary situations they encounter every day.
These assumptions include judgments about the physical properties, purpose, intentions
and behavior of people and objects, as well as possible outcomes of their actions and interactions.''

Inevitably, performing CSR requires having access to ``commonsense knowledge'' described in Wikipedia 
\cite{ref_wikipedia_2} 
as ``In artificial intelligence research, commonsense knowledge consists of facts about the everyday world, 
such as 'Lemons are sour', that all humans are expected to know. 
It is currently an unsolved problem in Artificial General Intelligence ...'' 

Due to its importance and a long-standing history, there is a large body of research devoted
to both designing algorithms and specialized logics for CSR as well as building commonsense knowledge bases.
While we have seen significant progress and popularity of AI in the last decade, mostly
due to progress in Machine Learning (ML), the progress in CSR has been slow and no major 
breakthroughs have arguably been achieved. Citing
\cite{ref_Davis_Marcus_1}: 
``AI has seen great advances of many kinds recently, but there is one critical area
where progress has been extremely slow: ordinary commonsense'' and O.Etzioni, 
the CEO of the Allen Institute for AI \cite{ref_etzioni}: ``No AI system currently deployed
can reliably answer a broad range of simple questions such as: If I put my socks in a drawer, 
will they still be in there tomorrow? or How can you tell a milk carton is full? 
For example, when AlphaGo beat the number one Go player in the world in 2016, 
the program did not know that Go is a board game.''

The current paper is driven by a goal to build hybrid systems which
complement the current machine learning systems for natural language with
logic-based reasoning systems specialized for CSR. 
We will present the GROK framework for extending the existing  algoritms and systems
of classical first order reasoning to CSR capabilities.

\section{Current approaches}

The specific CSR task targeted by the current paper is {\em question answering}: given 
either a knowledge base of facts and rules or a large corpora of texts (or both), plus
optionally a situation description (assumptions) for the questions,
answer questions posed either in logic or natural language. A simple often-used example
with inherently given choices is: given the assumption 
``The trophy would not fit in the brown suitcase because it
was too big'', answer ``What was too big?''. An example of an {\em open-ended question}
would be ``Which birds cannot fly?''.

The current approaches to CSR can be broadly classified as based on either logical reasoning
or machine learning (ML) on a large corpora of natural language texts. 

The main successes have been achieved relatively recently with the ML systems targeting
question answering in natural language (more details are given in the later benchmarks
chapter). A core method for ML approaches is to build a prediction system using
ML techniques, which, given an input text, predicts the next or missing word or phrase. 
The most prominent current underlying system is BERT \cite{ref_devlin_1} from Google AI,
with a number of specialised systems built on BERT.

Historically, the longest-going CSR project has been the logic-based CYC project, already in 1985
\cite{ref_lenat_1} stating the focus on CSR. The CYC project has created a largest known
manually made commonsense knowledge base along with reasoners specially built for this
knowledge base. Parts of the knowledge base -- OpenCyc -- are available as a large FOL axiom set 
\cite{ref_cyc} in
the TPTP \cite{ref_TPTP2} problem set for automated reasoning. 
Despite several successes, the approach
taken in the CYC project has often been viewed as problematic (\cite{ref_suguraman_1},
\cite{ref_domingos_1}) and has been repeatedly used  as an argument against logic-based
methods in CSR.

We note that the
CYC deduction problems in TPTP remain relatively hard for top-of-the-line automated
reasoners, mostly due to little focus on CYC-specific proof search heuristics.
However, the paper \cite{ref_hoder} introduced the SInE method for handling large theories,
which has been later implemented by several automated reasoning systems and significantly
enhances their efficiency for large knowledge bases. A recent paper \cite{ref_suda} 
reports important experimental gains from combining SInE with the methods from \cite{ref_tammet_2}.
For a concrete architecture for reasoning with large theories see \cite{ref_sutcliffe_spass}.
A overview of methods geared towards the similar goal is given in the paper \cite{ref_blanchette}.

CYC is not the sole current system for logic-based CSR. The papers \cite{ref_furbach_1}
\cite{ref_furbach_2} and \cite{ref_furbach_3} describe research and experiments
with a system for natural language question answering, converting
natural language sentences to logic and then performing proof search, using different
existing FOL knowledge bases. The authors note a number of difficulties, with
the most crucial being the lack of sufficiently rich FOL knowledge bases.

Due to the arguably shallow nature of current pure ML approaches, there is growing
interest in building hybrid systems, composing both ML and symbolic reasoning 
systems, both in the natural language and image recognition areas: see the
paper \cite{ref_marcus_3} for an overview of the current work in the area.
``Symbolic reasoning'' in this context is a vague term, sometimes, but
not always, meaning ``logic-based systems''. The paper \cite{ref_he_1} describes
a system obtaining new state-of-the-art results on three classic commonsense reasoning tasks.
It consists of two component models, a masked language model and a deep semantic similarity model
using unsupervised deep structured semantic models of \cite{ref_wang_2},
thus employing symbolic reasoning only in a fairly shallow way. In the last
paper the authors note that the previous example question
``The trophy would not fit in the brown suitcase because it
was too big. What was too big?'' still cannot be satisfactorily answered.

\section{Benchmarks for Question Answering}

An important question for research in CSR is the issue of measurable benchmarks.
The main benchmark set TPTP  \cite{ref_TPTP2} for Automated Reasoning (AR)
contains a relatively small subcategory for CSR problems in FOL.

Largely due to the progress in Natural Language Processing (NLP) achieved by ML
systems, a number of natural language question answering benchmarks
have been built and proposed in recent years. There is significant controversy 
around the ``gameability'' and superficial methods being sufficient for performing well
on some of these benchmarks. Newer benchmarks attempt to focus on questions which
require deeper understanding.

\cite{ref_wang_1} presents the SuperGlue benchmark, parts of which target
question answering. An example with a binary choice question:
``Premise: My body cast a shadow over the grass. Question: What’s the CAUSE for this?
Alternative 1: The sun was rising. Alternative 2: The grass was cut.´´. For these
kinds of questions the authors note that the best of their systems, BERT++, achieved
ca 74\% accuracy, while people achieved 100\%.

\cite{ref_clark_1} presents the ARC benchmark containing grade-school level, 
multiple-choice science questions, assembled to encourage research in advanced question-answering.
The current leader of the ARC public leaderboard is the
FreeLB-RoBERTa (single model) system, with the accuracy 68\% and several
BERT-based systems following closely.

\cite{ref_mihaylov} presents a natural-language based
dataset OpenBookQA for Open Book Question Answering. 
An example question from OpenBookQA:
``Which of these would let the most heat travel through?''
\begin{itemize}
\item a new pair of jeans.
\item a steel spoon in a cafeteria.
\item a cotton candy at a store.
\item a calvin klein cotton hat.
\end{itemize}
Answering the question requires using the given science fact ``Metal is a thermal conductor''
and common knowledge expected to be present or derived from the knowledge base of the
measured system: ``Steel is made of metal. Heat travels through a thermal conductor.''
The current public leaderboard for OpenBookQA has the system AristoRoBERTaV7
of the authors achieving best performance: 78\% accuracy.

\cite{ref_Marcus_2} presents a new natural language benchmark set and evaluates it for 
five top-of-the-line language models for question answering. As an example we bring two
challenges from the set, for which none of the tested systems managed to give a satisfactory answer:
\begin{itemize}
\item ``There are six frogs on the log. Two leave, but three join. The number of frogs on the log is now? 
Answer: seven.´'
\item ``Yesterday I dropped my clothes at the dry cleaners and have yet to pick them off. Where
are my clothes? Answer: at the dry cleaners.''
\end{itemize}
\cite{ref_Marcus_2} summarizes the findings thus: ``large-scale language models do a good job
of figuring the topic under consideration, and
what the plausible set of masked words / continuations might be, given the input context.
But a poor job of reasoning about which specific response is the right one.''

\section{Integrating machine learning for NLP with logic-based reasoning}

The hybrid question-answering algorithm we strive towards will
\begin{itemize}
  \item Use common ML-based methods for 
  \begin{itemize}
    \item determining the context of the question in the form of a set of tag/relevance measure pairs,
    \item parsing the input assumptions and the question to a logic-based formal representation,
    \item proposing a set of candidate answers along with their confidence measures.
  \end{itemize}  
  \item Use the common sense reasoning engine and a common sense knowledge base (CSKB) for
  \begin{itemize}
    \item updating the confidence of candidate answers, 
    \item proposing new candidate answers along with their confidence measures,
    \item providing derivations of the answers/confidences.
  \end {itemize} 
  \item Use a special converter to present the logic-based derivations in natural language as explanations of answers given.
\end{itemize}

The common sense reasoner should use the context tag/relevance measures provided by the ML 
component for
\begin{itemize}
\item updating the confidence measures of clauses in CSKB for the context of the questions (this would be similar, but more
fine-grained than CYC topics, see \cite{ref_cyc}),
\item heuristic preference of clauses likely relevant to the question (see \cite{ref_furbach_3} for relevant research).
\end{itemize}
 
The crucial questions out of the scope of this paper are: (a) how to represent knowledge in the CSKB, 
(b) how to convert natural language sentences to a logic-based formalism and 
(c) how to build a suitable CSKB. These questions are interrelated and have been an area of active research
for decades. 

One of the original goals of the CYC project \cite{ref_lenat_1} was automating knowledge acquisition from
natural language texts in order to build a comprehensive, ever-growing CSKB. This turned out to be harder than expected,
thus the CYC base was written mostly by people, arguably leading to brittleness. Given the recent successes of ML in NLP,
the possibility of automating CSKB building from, say, Wikipedia texts, has been under consideration by several research
groups.

There are several currently available large knowledge bases built for and used by both NLP and CSR research, like
WordNet, ConceptNet, NELL, Yago, DBpedia and OpenCyc. They use different formalisms and have widely different aims and
contents. Our initial investigations of these knowledge bases indicate that they do not, for the large part, contain
the most basic knowledge we would expect a small child to know. For example, we have not found
evidence from these knowledge bases for deriving classical default reasoning examples that objects 
typically do not fly, but birds fly, while dead birds and penguins do not fly. Similarly, basic information about the human
relatives seems to be mostly missing. Instead, the bulk of information is comprised of either detailed factual knowledge about actual people,
places and events or a semi-random selection of simple ontology rules for a very wide variety of concepts. Clearly,
the task of building a CSKB covering a significant percentage of the knowledge of a small child is still open.

We consider it likely that it is extremely hard to build a logic-based CSKB without having access to a practically usable and efficient
common sense reasoning engine: a chicken-and-egg problem.

\section{GROK extension framework for reasoning with CSR}

In the following we propose the GROK (Graph Reasoning On commonsense Knowledge) 
framework of extensions to the mainstream resolution-based search methods to make 
these capable of performing search tasks for practical commonsense reasoning with 
reasonable efficiency. We expect that the same framework can be adapted to search
methods different from resolution, 

The intuition behind GROK is preserving first order classical logic (FOL) intact
as an underlying machinery for derivations in CSR. The core methods of
automated reasoning used by most of the high-performance automated reasoning
systems (in particular, the resolution method) remain usable as core
methods for CSR. Essentially, FOL with the resolution method produces all
combinations of derivable true sentences modulo obvious simplifications
like subsumption. The main difference from strict FOL and GROK extensions
stands in the handling of constructed proof trees, which are evaluated
using various heuristic functions, estimating their relevance
and confidence. The outcome of a GROK reasoner is a set of FOL
proofs with these measures added. The actual use of measures in applications boils
down to the criteria for accepting or not accepting the derivations with measures under
the application-specific limits. 

While the proposed framework is focused on efficient proof search, we will not
discuss or propose useful modifications to algorithms and heuristics of the resolution
method crucial for efficiency, like subsumption, clause ordering and clause selection.
The specialization of these algorithms for CSR reasoning is left out
of the scope of the paper due to size limits.

\subsection{Resolution method}

In the following we will assume that the underlying first order reasoner uses the resolution
method, see \cite{ref_resolution_1} for details or Wikipedia page \cite{ref_resolution_2} 
for a quick overview. The rest of the paper assumes familiarity with the main concepts,
terminology and algorithms of the resolution method.

We treat the resolution method
as a combinatorial search algorithm providing all possible derivations for the questions posed.
On top of these we will add additional filters and measures, eliminating some derivations and
calculating confidences and signs for answers by additional operations on the derivations found.

\subsection{Queries and answers}

We assume the question posed is in one of two forms: 
\begin{itemize}
\item Is the statement $Q$ true?
\item Find values $V$ for existentially bound variables in $Q$ so that $Q$ is true.
\end{itemize}

For simplicity's sake we will assume that the statement $Q$ is in the prefix form, i.e. no quantifiers occur in
the scope of other logical connectives.

In the second case there could be several different assignment vectors to the variables,
essentially giving different answers. We also note that the answer could be a disjunction,
giving possible options, but not a single definite answer. However, as shown in \cite{ref_tammet_1},
in case a single definite answer exists, it will be derived.

A widely used machinery in resolution theorem provers for extracting values of existentially bound 
variables in $Q$ is to use a special {\em answer predicate}, converting a question
statement $Q$ to a formula 
$$\exists X_1,...,\exists X_n (Q(X_1,...,X_n) \& \neg answer(X_1,...,X_n))$$
for existentially quantified variables in $Q$. Whenever a clause is derived
which consists of only answer predicates, it is treated as a contradiction (essentially, answer)
and the arguments of the answer predicate are returned as the values looked for. 
A common convention is to call such clauses {\em answer clauses}.
We will require that the proof search does not stop whenever an answer clause is found, but will continue
to look for new answer clauses until the predetermined time limit is reached. Generally there
is no guarantee that all the answer clauses (or any answer clause) will be found in the given time limit.
See \cite{ref_multiple} for a framework of extracting multiple answers.

We also assume the queries take a general form
$$(CSKB \& A) \Rightarrow Q$$
where $CSKB$ is a commonsense knowledge base, $A$ is an optional set of precondition statements for
this particular question and $Q$ is a question statement. Despite the fact that one could use
a complex question statement $A \Rightarrow Q$ instead of a separate precondition and query
statement, or consider $A$ to be a part of $CSKB$, it is useful to distinguish these. We note
that the same distinction is made in a significant proportion of FOL problems in TPTP \cite{ref_TPTP}.

Since we assume the use of the resolution method for proof search, the whole general query
form is negated and converted to clauses, i.e. disjunctions of literals (positive or negative
atoms). We will call the clauses stemming from the question statement {\em question clauses}.

For better readability
we will present most of the disjunctions as implications.  We will use
single upper-case letters for universally quantified variables and words starting with 
lower-case letters for constants and predicate symbols.

The following is a simple example of the assumed structure, where the first two clauses
comprise the CSKB, the third to fifth clauses stem from the precondition statement and the
last clause is a question clause stemming 
from the question statement $\exists X fast(X)$ with the answer predicate added. 
\begin{verbatim}
bird(X) => canfly(X)
canfly(X) => fast(X)
bird(tweety)
bird(fleepy)
bird(creepy) V bird(bleepy)
fast(X) => answer(X)
\end{verbatim}

Three different answer clauses can be derived from this example: 
\begin{verbatim}
answer(tweety)
answer(fleepy)
answer(creepy) V answer(bleepy)
\end{verbatim}

\subsection{Inconsistencies}

We expect any CSKB with a nontrivial structure to contain inconsistencies in the sense
that a contradiction can be derived from the CSKB. Looking at existing CSKB-s mentioned
earlier, we observe that either they are already inconsistent (for example, the FOL version of 
OpenCyc \cite{ref_cyc} in TPTP \cite{ref_TPTP} is inconsistent) or would become inconsistent
in case intuitively valid inequalities are added, for example, inequalities of classes
like ``a cat is not a dog'', ``a male is not a female'' or default rules like ``birds can fly'',
``dead birds cannot fly'', ``penguins cannot fly''. We note that several large existing CSKB-s do
not contain such inequalities explicitly, although they are necessary for nontrivial question
answering.

Since classical FOL allows to derive anything from a contradiction, it is clearly unsuitable
for serious CSKB-s. There are two possible ways of overcoming the issue: either (a) use some version
of relevance logic \cite{ref_dunn_1} or other paraconsistent logics or 
(b) define a filter for eliminating irrelevant classical proofs.
We argue that despite a lot of theoretical work in the area, little work has been done in actual
automated proving for relevance logic, thus using
it directly is likely to create significant complexities. Instead we introduce a simple relevance
filter:

\begin{definition}[Relevance filter]
A resolution derivation of a contradiction not containing any answer clauses is discarded.
\end{definition}

Since a standard resolution derivation of a contradiction does not lead to any further derivations,
this filter is completeness preserving in the sense of finding all resolution derivations
containing an answer clause.

\subsection{Confidence}

Most, if not all of the commonsense knowledge is uncertain and in most cases uncertainty cannot
be adequately measured in practice. At the same time our practical experience gives us an intuitive
rough understanding of the distribution of these uncertainties.

Reasoning under uncertainty has been thoroughly investigated for at least a century, leading to a proliferation
of different theories and mechanisms (see, for example \cite{ref_richardson_1} and \cite{ref_braz_1}
), each of which is well suited for certain kinds of problems
and ill-suited for other kinds. 
The underlying reason of proliferation is the philosophical complexity of 
interpreting probability: see \cite{ref_hayek_1} for an overview. We claim that there is no 
universal and widely accepted theory or methodology for handling uncertainty in reasoning. 
Davis argues in \cite{ref_davis_1} that probabilistic models are not suitable for 
automated reasoning or plausible for cognitive models. 
We also note that ML systems use confidence measures pervasively
in the internal calculations and output results, yet most researchers avoid associating these measures with any
concrete philosophical approach or mathematical theory. Yann LeCun in \cite{ref_lecun_1} says 
``I am perfectly ready to throw probability theory under a bus''.

Philosophically we largely follow the {\em subjective interpretation} of probability as a degree of
belief, originating from Ramsey and later systematized by Bruno De Finetti. 
We use the word {\em confidence} to denote
our rough adherence to this interpretation. We avoid using complex measures like intervals,
distributions or fuzzy functions. 

More concretely, we take the approach of (a) providing a simple sensible baseline algorithm 
and (b) leaving open ways to modify this algorithm for specific cases as need arises.

We will use a single rational number in the range $0...1$ as a measure of a confidence of a clause,
with $1$ standing for perfect confidence and $0$ standing for no information. Confidence of a 
clause not holding is encoded as a confidence of the negation of the clause. For any given
clause $C$ we could, for example, ask a number of people a question 
``what is your degree of belief that $C$ holds, from $-1$ for certainty that $C$ does not hold to $1$
for certainty that $C$ does hold'', calculate the average answer $c$ and store the absolute
value of $c$ as a confidence of $C$ or $\neg C$, depending on the polarity of $c$. 
In practice we expect that an ML system employed in the construction of a CSKB would provide
us confidence measures without any clear indication of how or why exactly these measures were calculated.

As a baseline we use the arguably simplest approach of computing uncertainties of derived clauses:
multiplying confidences of statements under assumption that they are independent, stemming
from the Bayes' theorem as $P(A \land B) = P(A) * P(B|A)$. 

\begin{definition}[Confidence calculation for resolution steps]
For binary resolution and paramodulation steps the confidence of a result is obtained
by multiplying the confidences of the premisses. For the factorization step the confidence
of the result is the confidence of the premise, unchanged. The question clauses have confidence 1.
\end{definition}

We can argue that possible dependence of the premisses could be taken into account, exactly
as in the following section for cumulative evidence. However, here we prefer to err on the conservative
side, i.e. assuming the independence of the premisses.

A simple example employing forward reasoning, concretely, negative ordered resolution:
\begin{verbatim}
bird(tweety): 0.8
bird(X) => canfly(X): 0.9  
canfly(X) => fast(X): 0.7
fast(X) => answer(X): 1
\end{verbatim}
leads to a sequential derivation of 
\begin{verbatim}
canfly(tweety): 0.72
fast(tweety): 0.504
answer(tweety): 0.504
\end{verbatim}

We will now look at the situation with additional evidence for the derived answer.
In our context, using additional evidence is possible if a clause $C$ can be 
derived in different ways, giving two different derivations $d_1$ and $d_2$ with confidences 
$c_1$ and $c_2$. In case the derivations $d_1$ and $d_2$ are independent, we could apply the 
Bayes theorem as $P(A \lor B) = P(A) + P(B) - P(A \land B)$ to calculate the cumulative confidence.  

What would it mean for derivations to be ``independent''? In the context of commonsense reasoning
we cannot expect to have an exact measure of independence. However, suppose the derivations
$d_1$ and $d_2$ consist of exactly the same initial clauses, but used in a different order.
In this case $c_1=c_2$ and the cumulative confidence should intuitively be also just $c_1$: no additional
evidence is provided. On the other hand, in case the non-question input clauses of $d_1$ are $d_2$ are 
mutually disjoint, then - assuming all the input clauses are mutually independent - the
derivations are also independent, and we should apply the previous rule for $P(A \lor B)$ for
computing the cumulative confidence.

We will estimate the independence $i$ of two derivations $d_1$ and $d_2$ simply as 
one minus the ratio of shared input clauses to the total number of input clauses in $d_1$ and $d_2$.
Thus, if no clauses are shared between $d_1$ and $d_2$, then $i=1$ and if all the clauses are shared, 
then $i=0$.

In addition, we also know that it is highly unlikely that all the input clauses 
are mutually independent. 
Again, lacking a realistic way to calculate the dependencies, we can 
simply give a heuristic estimate $h$ in the range $0...1$ to the overall independence
of the input clause set, where $1$ stands for total independence and $0$ for total
dependence. The heuristic estimate can be viewed as our measure of optimism about the 
overall indepedence.

Finally, we will calculate the overall independence of two derivations $d_1$ and $d_2$ as
$i*h$. We will use the combination of these two independence measures in the following rule
in a somewhat ad-hoc manner as follows.

\begin{definition}[Confidence calculation for cumulative evidence]
Given two derivations $d_1$ and $d_2$ of the search result $C$ with confidences $c_1$ and $c_2$,
calculate the updated confidence of $C$ as 
$$max(c_1 + c_2*i*h - (c_1*c_2*i*h), c_1*i*h + c_2 - (c_1*c_2*i*h))$$
where
\begin{itemize}
\item $i$ is one minus the number of shared non-question input clauses in $d_1$ and $d_2$ divided by the total number of
non-question input clauses in  $d_1$ and $d_2$,
\item $h$ is the heuristic estimate of the independence of the total set of input clauses from $1$ for
total independence to $0$ for total dependence.
\end{itemize}
\end{definition}

Despite the ad hoc nature of the usage of $i$ and $h$, the formula satisfies the
following {\em intuitive requirements for cumulative evidence}:
\begin{itemize}
\item If $d_1$ and $d_2$ do not share non-question input clauses and the
total set of input clauses is mutually independent, $i*h=1$ and the formula degrades to
$c_1 + c_2 - (c_1*c_2)$.
\item If $d_1$ and $d_2$ have the same non-question input clauses or the
total set of input clauses is mutually totally dependent, $i*h=0$ and the formula degrades to 
$max(c_1,c_2)$.
\end{itemize}

As an example, let us consider new input clauses and another way to derive the same answer:

\begin{verbatim}
inair(tweety): 0.8
inair(X) => fast(X): 0.9   
fast(X) => answer(X): 1
\end{verbatim}
leads to a sequential derivation of 
\begin{verbatim}
fast(tweety): 0.72
answer(tweety): 0.72
\end{verbatim}

We observe that the sets of non-question clauses
in the derivations are different, thus $i=1$. 
If the total heuristic independence measure $h=1$, we get the cumulative
confidence $0.504 + 0.72 - (0.504*0.72)=0.86112$. Setting $h=0.5$ leads to
\begin{eqnarray*}
max(0.504 + 0.72*0.5 - (0.504*0.72*0.5), \\ 0.504*0.5 + 0.72 - (0.504*0.72*0.5)) & = & \\
max(0.68256,0.79056) & = & \\
0.79056
\end{eqnarray*}
We note that one could use a different formula for calculating the cumulative confidence.
For example, the confidences of clauses could be taken into account while computing the
independence measure $i$. However, we argue that the previously given intuitive 
requirements should be followed.

Next, consider the question of whether we should calculate the cumulative confidence
for all the clauses derived, not just the answer clauses, as done in our framework.
It is an interesting separate problem to determine under which conditions the results for the answer clauses
would be same for both approaches, and we will not tackle it here. However, we argue that
\begin{itemize}
\item the inexactness of confidence measures is likely to dominate over the differences possibly surfacing
for different approaches or for minor modifications of the given formula,
\item it is more efficient to calculate the cumulative confidence of the answer clauses
as opposed to calculating these for all the clauses in the derivation: the latter would both require additional search
and would raise the question as for how to cumulate confidences of inequal, but subsuming clauses.
\end{itemize}

Observe that neither the overall set
nor the structure of derivations depends on the confidence measures; the forthcoming
parts of our framework keep this property. Suppose that we have two different derivations $d_1$ and
$d_2$ which lead to the same answer, but share an intermediate derived clause $C$. In our framework
the confidence measure of $C$ is used for computing the confidence of the answers before the
cumulative evidence rule is used. 

\subsection{Collecting negative evidence}

Since we assume that the CSKB can be inconsistent and we use the previously described 
relevance filter for eliminating derivations not containing clauses stemming from the
query, we must consider the possibility that both the query and the negation of the
query could be derivable.

Thus, we conduct additional proof search for the negations of the previously assumed 
two types of questions $Q$:

\begin{itemize}
\item If $Q$ contains no existentially quantified variables, is the statement $\neg Q$ true?
\item For all $i$ vectors of values $C_1i,...,C_ni$ found for existentially bound variables 
$X_1,...X_n$ in $Q$ making $Q$ true, is $\neg Q$ true when we substitute the values in $C_1i,...,C_ni$ 
for corresponding variables in $Q$?
\end{itemize}

The proof tasks are created by clausifying the corresponding negated and substituted-into questions
along with the CSKB and precondition statements. The overall algorithm for question answering 
is as follows:

Assuming we have a given amount of time $t$ and the coefficient $c$ in $0...1$
for the proof search for $Q$, conduct the search as follows:
\begin{enumerate}
\item Spend $t*c$ for proof search for $Q$, possibly generating several derivations with $i$ different value vectors for 
existentially quantified variables in $Q$.
\item Compute the cumulative confidences for answers with $i$ different value vectors.
\item Spend $t*(1-c)$ for finding answers to $\neg Q(C_1i,...,C_ni)$, iterating over $i$ different value vectors 
found in the first step. Each iteration is a full stand-alone resolution search possibly leading to several
derivations.
\item Compute the cumulative confidences for answers with different negative value vectors collected in
the last step.
\item Compute the sign and the overall confidence for different value vectors found in the first step.
\item For the existential-variable-free question $Q$ return the answer {\em true} or {\em false} along with the
overall confidence and relevant derivation trees. For the case with variables, return only positive
answers along with the confidence and relevant derivation trees.
\end{enumerate}

For sign and overall confidence calculation we employ the following simple rule subtracting the negative
confidence from the positive confidence:

\begin{definition}[Sign and confidence calculation for combined positive and negative evidence]
Given a question $Q$ and the vector of values $C_1,...,C_n$ for the existentially quantified variables
in $Q$, calculate $cp-cn$ where $cp$ is the cumulative confidence for the positive answer and $cn$ 
is the cumulative confidence for the negative answer. If $cp-cn<0$, the answer is negative with the confidence
$0-(cp-cn)$, otherwise the answer is positive with the confidence $cp-cn$.
\end{definition}

\subsection{Default rules}

In addition to the observation that most of the common sense rules and facts are uncertain,
we also know that a large subset of these rules can be considered to be ``default'':
in the absence of contradictory evidence to the result of the application of the rule
we should assume the rule is true, while contradictory evidence should block the
application of the rule. A famous ``penguin example'' is as follows:

\begin{verbatim}
bird(tweety)
penguin(pengu)
penguin(X) => bird(X)
bird(X) => canfly(X)
penguin(X) => -canfly(X)
\end{verbatim}

We should use the bird-can-fly rule to derive that tweety can fly, while we should not use the same
rule to derive that pengu can fly, although we have penguin-is-a-bird rule. Moreover, we intuitively
expect that we should derive that penguin cannot fly, despite the obvious contradiction.

Another famous example is the ``Nixon triangle'', the most common so-called ``cautious''
interpretation of which is that we should neither derive that Nixon is a pacifist nor that
he is not a pacifist:

\begin{verbatim}
quaker(nixon)
republican(nixon)
quaker(X) => pacifist(X)
republican(X) => -pacifist(X)
\end{verbatim}

Default logic, proposed by R.Reiter in \cite{ref_reiter_1} has been one of the leading
and most researched contenders for handling default rules. Default logic,
along with autoepistemic logic and McCarthy's circumscription method belongs
to a family of nonmonotonic logics \cite{ref_brewka_1}: differently from classical logic,
a derivation in non-monotonic logic can block or invalidate another derivation. A similar
approach is taken with a Closed World Assumption where a negation $\neg A$ of any not derivable
statement $A$ is considered to be true.

The fundamental difference from the confidences analyzed before is that such confidence calculations
cannot normally capture the intuitive expectations we have with the default rules. For example,
we have no grounds to state that the bird-can-fly rule has much higher or much lower confidence
than the penguin-cannot-fly rule. Additionally we observe that the default rules are often associated
with multi-level hiearchies like
\begin{verbatim}
penguin(X) => bird(X)
bird => materialobject(X)
materialobject => -canfly(X)
bird(X) => canfly(X)
penguin(X) => -canfly(X)
\end{verbatim}

There are two major problems in the application of default rules. First, how to prioritize rules, second,
how to perform search efficiently. A naive algorithm would have to perform full, potentially nonterminating
sub-search for a blocking contradiction for each derivation step. 
So far the practical reasoning systems dealing with default logic have originated from the
answer set programming paradigm, 
employing propositional solvers on top of grounded first order logic formulas: 
see the overview paper \cite{ref_niemela}.

We will follow an important observation made in a number of papers concerning default logic see
e.g. \cite{ref_brewka_2} and \cite{ref_baader_1}: 
we should prioritize the more specific rules over more general rules. Here the ``generality'' is 
normally meant to indicate the layer of a taxonomoy hierarchy to which the rule is applicable. In the example
above the first two rules are absolutely certain {\em taxonomy rules} defining more and more 
abstract concepts penguin - bird - materialobject, not uncertain observational rules.
The underlying intuition is that our confidence that a certain bird can fly should be considered
not as a fixed number, but strongly dependent on what kind of bird we have. Statistically it is indeed
true that almost all (alive) birds can fly, while it is also true that none or almost none of the penguins
can fly. The crux of the observation is that the small set of non-flying birds is not evenly spread over
all kinds of birds, but heavily represented with penguins. 

The approach we take regarding default rules is to identify and annotate the taxonomy rules
in the CSKB as such and then use the intuition above to discard derivations where
a more general default rule application is contradicted with a less general application. 
Our current intuition is that all the rules with confidence below 1 should be treated
as default rules. 

We tackle the efficiency problem by not performing sub-searches for blockers. Instead, we check each
derivation of the answer for blockers, expecting that derivable blockers may have been derived during search
and be already present in the set of derived clauses. The more time we give for search, the more
likely the latter is to happen. In other words, more time given to a search may result in fewer
unblocked derivations of answers.

The following rough ad-hoc algorithm captures both of the basic intuitions described above.  Since we do not perform blocking
during search, when given unbounded time and resources, all the derivations, hence also all blockers, will be found. 
The algorithm is conservative in the following sense: it may block more derivations than we would intuitively like. It is
an important research task to investigate further improvements to the algorithm to make it less conservative; in particular,
instead of simply counting the number of taxonomy rule applications to measure specificity, we could perform a more fine-grained
comparison.

\begin{definition} [Default elimination algorithm]
Treat the derivation trees as directed acyclic graphs where each clause occurs only once.
Observe that all the clauses derived during search are sorted by their order of derivation and clauses derived later cannot occur 
in the derivations of earlier clauses. Let $d$ be a derivation of an answer clause. By a ``direct contradiction derivation of $x$'' we
mean derivations with exactly $n$ steps where $n$ is the number of non-answer literals in $x$: the motivation for looking for these is search
speed.
\begin{itemize} 
\item Mark all the clauses derived during search as unblocked.
\item Iteratively apply the following blocking check over all the clauses $C_i$ in the derivation $d$, starting from the earlier clauses:
\begin{itemize}
\item Determine whether one of the parent clauses of $C_i$ is a default rule. 
\item If yes, search for and loop over all the direct contradiction derivations $dc$ for $C_i$ and some of the already derived and 
unblocked clauses in the search space:
\begin{itemize}
\item If yes, determine whether the derivation $dc$ of the direct contradiction $C_i$ contains a smaller number of taxonomy rules than the
derivation of $C_i$.
\item If yes, determine whether the derivation $dc$ cannot itself be invalidated by the default elimination algorithm,
\item If it can, mark $dc$ and all the derivations containing $dc$ blocked, otherwise mark $C_i$ and all the derivations containing $C_i$ as 
blocked and break out of the current loop.
\end{itemize}
\end{itemize}
\item Determine whether the final answer clause is blocked and if yes, discard the derivation.
\end{itemize}
\end{definition}

\subsection{Similarities}

An important part of CSR is reasoning by analogues and similarities. Similarity-based reasoning in FOL has been
researched before, see e.g. \cite{ref_formato_1} and \cite{ref_weber_1}. A recent paper \cite{ref_jakubuv} 
describes an implementation of a neural guidance system for saturation-style automated theorem provers.

In the simplest cases we incorporate similarity-based reasoning by treating 
similar predicates/functions/constants as
equivalent or equal with a measure of confidence approximated from the given similarity measure. For example, given that kings are similar
to queens with a measure $s$ we create an implicit axiom {\tt king(X) <=> queen(X)}, the  confidence of which is calculated 
by a heuristic algoritm from $s$. This allows the reasoner to produce derivations carrying the knowledge associated with one kind of object
over to similar kinds, but with diminished confidence. 

In case we additionally know that {\tt king(X) => male(X)} and {\tt queen(X) => female(X) } with confidence 1, we apply algorithms similar to
the default reasoning described in the previous section. We will skip the details.

\section{Summary and further work}

We have presented a GROK framework for performing logic-based common sense reasoning by building upon existing 
algorithms and systems for automated reasoning in classical first order logic. We argue that such symbolic AI methods
have a good perspective to complement machine learning systems and achieve both deeper reasoning and explanatory capabilities
than is possible with the pure neural computation approach.

The described framework is currently in the implementation phase as a system GK, 
containing a large set of extensions and modifications to the existing automated reasoning system GKC
of the author: \cite{ref_tammet_2}, \cite{ref_gkc}. The efficiency-focused modifications to the common
algorithms and heuristics like subsumption, literal and clause selection in the CSR context will be 
covered in future papers.

We see both the practical experiments with the reasoners and the buildup and integration of commonsense knowledge bases with the
help of machine learning systems as the main driver of future research. The GROK framework covers a number of important, well-known
themes in AI, all of which can and should be investigated further from the practical standpoint of creating systems 
enhancing the quality and performance of NLP question answering.


%
%
%

\end{document}